\documentclass[sensors,article,accept,moreauthors,pdftex,10pt,a4paper]{mdpi} 

 \graphicspath{ {}} 
\usepackage{cite}
\usepackage{array}
\usepackage{color}
\usepackage{soul}
\usepackage{array,multirow,graphicx}
\usepackage{tabularx}
\usepackage{hyperref}
\usepackage{graphicx}
\usepackage{caption}
\usepackage[labelformat=simple]{subcaption}

\DeclareCaptionLabelFormat{subcaptionlabel}{\normalfont(\textbf{#2}\normalfont)}
\usepackage{amsmath}
\usepackage{algorithm}
\usepackage[noend]{algpseudocode}
\usepackage[utf8]{inputenc}
\usepackage{color,soul}
\usepackage{environ}
\NewEnviron{myequation1}{%
\begin{equation}
\scalebox{0.85}{$\BODY$}
\end{equation}
} 
\PassOptionsToPackage{hyphens}{url}\usepackage{hyperref}

\def\UrlBreaks{\do\/\do-}
\firstpage{1} 
\articlenumber{894}
\doinum{10.3390/s18030894}
\pubvolume{18}
\pubyear{2018}
\copyrightyear{2018}
\externaleditor{ }
\history{Received: 6 February 2018; Accepted: 15 March 2018; Published: 17 March 2018}

\usepackage{booktabs} 
\usepackage{multirow}
\usepackage{soul} 
\usepackage{microtype}
\setitemize{parsep=6pt,itemsep=0pt,leftmargin=*,labelsep=5.5mm}
\setenumerate{parsep=6pt,itemsep=0pt,leftmargin=*,labelsep=5.5mm}
\setlist[description]{itemsep=0mm} 

\Title{Fusion of an Ensemble of Augmented Image Detectors for Robust Object Detection}

\Author{Pan Wei $^{1}$,  John E. Ball $^{1,}$*  
and Derek T. Anderson $^{2}$} 

\AuthorNames{Pan Wei, John Ball, and Derek Anderson}

\address{%
$^{1}$ \quad Mississippi State University, Electrical and Computer Engineering Department, Starkville, MS 39759, USA; pw541@msstate.edu \\
$^{2}$ \quad University of Missouri-Columbia, Electrical Engineering and Computer Science Department, Columbia, MO~65211, USA; andersondt@missouri.edu } 

\corres{Correspondence: jeball@ece.msstate.edu; Tel.: +1-662-325-4169}

\abstract{A significant challenge in object detection is accurate identification of an object's position in image space, whereas one algorithm with one set of parameters is usually not enough, and the fusion of multiple algorithms and/or parameters can lead to more robust results. Herein, a new computational intelligence fusion approach based on the dynamic analysis of agreement among object detection outputs is proposed. Furthermore, we propose an online versus just in training image augmentation strategy. Experiments comparing the results both with and without fusion are presented. We demonstrate that the augmented and fused combination results are the best, with respect to higher accuracy rates and reduction of outlier influences. The approach is demonstrated in the context of cone, pedestrian and box detection for Advanced Driver Assistance Systems (ADAS)~applications.}

\begin{document}
\section{Introduction} 
\label{sect:intro}

Object detection is an important and active area in machine learning research. In many instances, especially for Advanced Driver Assistance Systems (ADAS), it is important to accurately localize objects in real time. 
Due to the development of deep learning, the localization accuracy for object detection has largely been improved. 
However, in our experimenting with deep learning for object detection (such as detecting pedestrians), we noticed that even with a well-trained network with image augmentation (such as translation, rotation, stretching) during training, changes in the image characteristics such as brightness or contrast can still have a large effect on the detection result. 
Based on these observations, we hypothesize that presenting a deep network with multiple augmented images during {\emph{testing}} and fusing the results could result in a more robust detection system. Herein, we develop a system utilizing image augmentation combined with Axis-Aligned Bounding Box Fuzzy Integral (AABBFI)-based fusion to enhance the detection results. The fusion method applied in this paper originates from the field of computational intelligence. This method analyses the agreement among object detection outputs and~fuses them dynamically. We choose not to use a supervised learning-based method (learn the fusion from training data), because the fusion method in this scenario cannot be learned in general, as the ``optimal'' fusion result changes from case to case. Specifically, even with the same inputs are to be fused, the expected fusion result is different when the environmental surroundings and/or detection object type changes. Instead, the method we apply in this paper automatically evaluates the characteristics among inputs for each case and fuses the detection results dynamically. 

The primary research task we were investigating for this paper is developing a robust detection sub-system for ADAS. During road construction, street repairs and accidents, often, traffic lanes are quarantined with barrels or cones. Furthermore, ADAS detect pedestrians to estimate their locations and help with collision avoidance. In our specific industrial project, we also need to detect boxes. This~motivated the study of cone, pedestrian and box detection herein, and a dataset consisting of traffic cones, pedestrians and boxes has been collected and analyzed. In the experimental results, we also include results using the PASCAL VOC (Visual Object Classes) dataset \citep{everingham2010pascal} for generalization~purposes. 

This paper shows that the proposed method gives better detection results than both the original non-fusion results and the average/median fusion results. 
Specifically, the contributions of this work~are:
\begin{itemize}
  \item A detection fusion system is proposed, which uses online image augmentation {\emph{before the detection stage}}  
  (i.e., during testing) and fusion after to increase detection accuracy and robustness. 
  \item The proposed fusion method is a computationally-intelligent approach, based on the dynamic analysis of agreement among inputs for each case. 
  \item The proposed system produces more accurate detection results in real time; therefore, it helps with developing an accurate and fast object detection sub-system for robust ADAS. 
  \item The Choquet Integral (ChI) is extended from a one-dimensional interval to a two-dimensional axis-aligned bounding box (AABB).
\end{itemize}

The following acronyms and mathematical nomenclature, shown in Table \ref{table:Notation}, are used in the paper. 
\begin{table}[H]
\centering
\caption{Acronyms and mathematical nomenclature}.  

\label{table:Notation}
\begin{tabular}{cc}
\toprule
\textbf{Variable/Symbol} & \multicolumn{1}{c}{\textbf{Description}}  \\ \midrule
AABB         & Axis-Aligned Bounding Box \\ 
AABBFI       & Axis-Aligned Bounding Box Fuzzy Integral \\ 
ADAS         & Advanced Driver Assistance Systems \\ 
ChI         & Choquet Integral \\ 
FI         & Fuzzy Integral \\ 
FM         & Fuzzy Measure \\ 
FPS         & Frames Per Second \\ 
IoU         & Intersection over Union \\ 
NMS         & Non-Maxima Suppression \\

$x_{i}$        & $i${-th} data/information input \\ 
$X=\{x_1,...,x_n\}$   & Finite set of $n$ data/information inputs \\ 
$I$          & Set of intervals, $\{[i^l, i^r]\in I, i^l \leq i^r, i^l, i^r \in \Re \}$\\ 
$g$          & Fuzzy measure $g$: $2^{X}$ $\to$ [0,1] \\ 
$g(\{x_i\})$      & Fuzzy measure on input $x_i$ \\ 
$g(\{x_i,..., x_j\})$  & Fuzzy measure on inputs $x_i,..., x_j$ \\ 
$h$          & Real-valued evidence function $h : X \rightarrow \Re$ \\ 
$h(x_i), h_i$    & Real-valued evidence from input $x_{i}$\\ 
$\bar{h}_{i}$     & Interval-valued evidence from input $x_{i}$ \\ 
$[\bar{h}_{i}]^{-}$   & Left endpoint of $\bar{h}_{i}$  \\ 
$[\bar{h}_{i}]^{+}$   & Right endpoint of $\bar{h}_{i}$  \\ 
$\int{h} \circ g$   & Fuzzy integral of $h$ with respect to $g$ \\
 
$h(x_{\pi(i)})$    &$\pi$ is a permutation of $X$, such that $h(x_{\pi(1)}) \geq h(x_{\pi(2)}),...,\geq h(x_{\pi(n)})$   \\ \bottomrule

\end{tabular}
\end{table}

This paper is organized as follows: Section \ref{sect:background} introduces basic concepts related to the Fuzzy Integral (FI). Furthermore, it introduces recent developments in object detection for ADAS, image augmentation in deep learning and model ensembles. Section \ref{sect:methods} discusses the proposed fusion system, and Section \ref{sect:results} gives synthetic and ADAS experimental examples showing how fusion and the system work. Finally, Section \ref{sect:conclusionsfuture} contains conclusions and future work.

\section{Background} 
\label{sect:background}

In the proposed detection fusion system, a fusion method from the field of computational intelligence is applied after the detection stage to increase the detection accuracy. 
There are various methods performing fusion in a computationally-intelligent way, and one example is in~\citep{wei2015measuring, wei2016multi}. 
Herein, the~Fuzzy Integral (FI), a parametric nonlinear aggregation operator, is used for fusion~\citep{sugeno1974theory, murofushi1989interpretation, grabisch2000fuzzy}.
To~facilitate a better understanding of this fusion method, in this section, the definitions of Fuzzy Measure (FM), Choquet Integral (ChI), interval-valued fuzzy integral and FM of agreement are reviewed. 
Next, object detection for ADAS, image augmentation in deep learning and model ensembles are discussed. 

  \subsection{Fuzzy Measure} 
  \label{subsect:FM}
  The ChI is defined with respect to the Fuzzy Measure (FM). The FM is the modeling of the worth of the individual inputs and their various interactions. Ultimately, once we select an FM, the ChI becomes a specific operator. 
  
  Let $X=\{x_1,...,x_n\}$ be a set of $n$ information inputs (e.g., experts, sensors, algorithms, etc.), where $x_i$ represents the $i${-th} input. For example, in our application, $x_i$ is the $i${-th} AABB. The definition of FM is as follows,

 \begin{Definition}[\bf{Fuzzy measure}]
   For a finite set $X$, a measure $g$: $2^{X}$ $\to$ [0,1] is an FM if it has the following~properties:
   \begin{enumerate}
   \item (Normality) $g(\emptyset) = 0$,
   \item (Monotonicity) If $A, B \in 2^{X}$ and $A \subseteq B \subseteq X$, then $g(A) \le g(B) \le 1$.
   \end{enumerate}
  \end{Definition}

 Note, often, $g(X)=1$ for problems like confidence/decision fusion. Property 2 of the FM states that $g$ is monotone.
  For example, if there are three inputs $\{x_1, x_2, x_3\}$, then the FM $g(\{x_1, x_2, x_3\})$ $\geq$ $max\left(g(\{x_1, x_2\}), g(\{x_1, x_3\}), g(\{x_2, x_3\})\right)$ and $g(\{x_1, x_2\})$ $\geq$ $max(g(\{x_1\}), g(\{x_2\}))$. What this means is adding inputs never decreases the measure. 
  
\begin{figure}[H]
  \centering
  \includegraphics[height=.25\textheight]{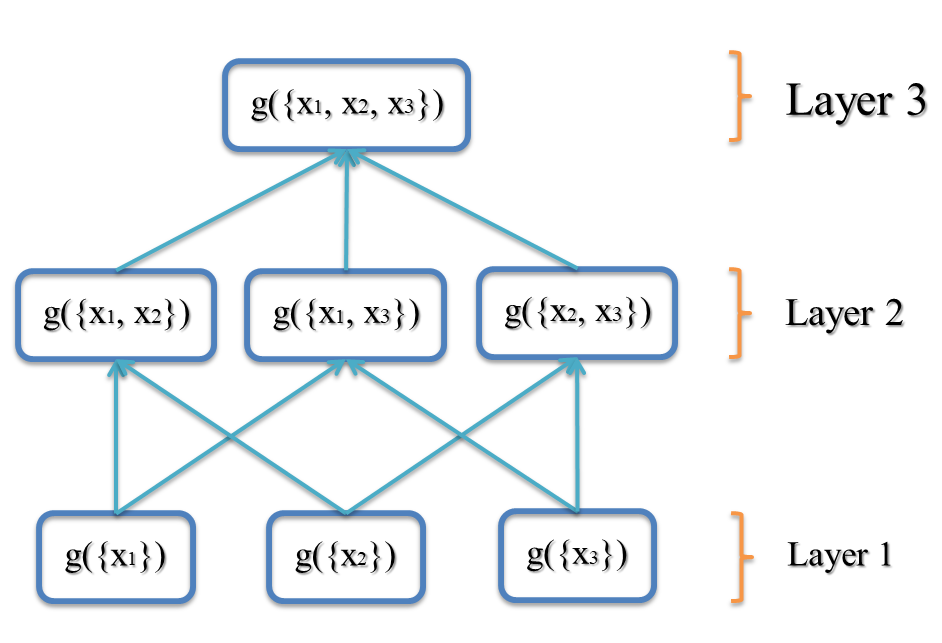}
  \caption{Fuzzy measure shown for three information inputs.}
  \label{fig:layer}
  \end{figure}

  The FM ($g$) lattice (a graph that shows the variables and the minimal set of required monotonicity constraints as edges) for three inputs ($N=3$) is shown in Figure~\ref{fig:layer}. 
  In the first layer, the variables are only for a single input. In the second layer, variables are for two inputs, and so forth. 

  \subsection{Choquet Integral} 
  \label{subsect:FI}
  
  Many variants of the fuzzy integral have been put forth to date \citep{grabisch2000fuzzy}, e.g., the Sugeno FI \citep{sugeno1974theory}, the~Choquet Integral (ChI) \citep{murofushi1989interpretation}, etc. Herein, we use the ChI, as it produces results anywhere between the minimum and maximum inputs. The Sugeno FI only yields one of the $N$ inputs or $2^N$ FM variable values due to its min/max formulation. 
  
  \begin{Definition}[\bf{Choquet integral}]
  Let $h(x_i)$ represent real-valued evidence from input $x_{i}$. Let $g(\{x_i,..., x_j\})$ represent the FM on inputs $x_i,..., x_j$. Let $\pi$ be a permutation of $X$, such that $h(x_{\pi(1)}) \geq h(x_{\pi(2)}),..., \geq h(x_{\pi(n)}) $, $A_{i} = \{x_{\pi(1)},..., x_{\pi(i)}\}$ and $g(A_{0}) = 0$. Let $h : X \rightarrow \Re$ be a real-valued function that represents inputs. The ChI is: 
  \begin{equation}
  \int_{c}h \circ g = C_{g}(h) = \sum_{i=1}^{n}h(x_{\pi(i)})[g(A_{i}) - g(A_{i-1})].
  \label{eq:CFI}
  \end{equation}
  An example showing this calculation is shown in Section \ref{subsect:se}.
  \end{Definition}
  
  \subsection{Interval-Valued Fuzzy Integral}
  In some cases, evidence $h$ is not represented as a single value, but as an interval. 
  Intervals capture uncertainty at a level of granularity that is acceptable for many applications. For example, a radar might measure range and give a value, but this value has uncertainty, so it could be represented as an~interval.
  
  \begin{Definition}[\bf{Interval-valued fuzzy integral}]
  Let $\bar{h}_{i} \subseteq I$ be the continuous interval-valued evidence from input $x_{i}$, and let $[\bar{h}_{i}]^{-}$ $\leq$ $[\bar{h}_{i}]^{+}$ be the left and right endpoints of interval $\bar{h}_{i}$, respectively. The interval-valued FI is \citep{grabisch2000fuzzy}:
  
  \begin{equation}
  \int\bar{h} \circ g = [\int[\bar{h}]^{-} \circ g , \int[\bar{h}]^{+} \circ g ].
  \label{eq:IFI}
  \end{equation}
 \end{Definition}
  
  This equation means that interval-valued input breaks down into the extreme (interval boundaries) case. This is the same for set-valued input, as well. In \citep{anderson2014extension}, there is a review of different extensions of the FI for uncertain $h$ and/or $g$. 

  \subsection{Fuzzy Measure of Agreement}
  \label{subsec:fma}
  The FM can be (i) specified by an expert, (ii) learned from the training data, (iii) input from densities or (iv) extracted from the current observation (this paper). 
  One example to extract the FM from the current observation is the FM of agreement proposed in \citep{wagner2012fuzzy, havens2013fuzzy, havens2014data}. 
  
  The motivation for the proposal of the FM of agreement is that sometimes, experts' estimation or the ground-truth can be challenging to obtain, and all we have is the data themselves. This means that we have the inputs, but we do not know their ``worth''. How the FM of agreement decides the ``worth'' of each input is by calculating the amount of agreement between each input and other inputs. Herein, agreement means the amount of overlap among different inputs. In the extreme case, if one input has no overlap with any other inputs, the FM of agreement for two-tuples is zero. 
  
  The following demonstrates the way to calculate the FM of agreement when the available evidence $\bar{h} = \{ \bar{h}_1, ... , \bar{h}_n \}$ is interval-valued.
  
  Let $\bar{A}_i = \{ \bar{h}_{\pi(1)}, ... , \bar{h}_{\pi(i)} \}$ be the permuted set of intervals and $z_i$ be the weight of each term, where $z_2 \le z_3 \le ... \le z_n$. In \citep{wagner2012fuzzy}, $z_i = i / n$ and:
\vspace{-12pt}

  \begin{subequations}
  \begin{align}
  \tilde{g}^{AG}(\bar{A}_0) 
  &= \tilde{g}^{AG}(\bar{A}_1) = 0, 
  \label{eq:MAa}\\
  \begin{split}
  \tilde{g}^{AG}(\bar{A}_i) 
  &= \bigg| \bigcup_{k_1=1}^{i-1} \bigcup_{k_2=k_1 + 1}^{i} \bar{h}_{\pi(k_1)} \cap \bar{h}_{\pi(k_2)} \bigg|z_2 \\
  &+ \bigg| \bigcup_{k_1=1}^{i-2} \bigcup_{k_2=k_1 + 1}^{i-1} \bigcup_{k_3=k_2 + 1}^{i} \bar{h}_{\pi(k_1)} \cap \bar{h}_{\pi(k_2)} \cap \bar{h}_{\pi(k_3)} \bigg|z_3 \\
  &+ ... + | \bar{h}_{\pi(1)} \cap \bar{h}_{\pi(2)} \cap ... \cap \bar{h}_{\pi(i)} | z_i,
  \end{split}
  \label{eq:MAb}
  \\
  g^{AG}(\bar{A}_i)
  &= \dfrac{\tilde{g}^{AG}(\bar{A}_i)}{\tilde{g}^{AG}(\bar{h})}, \quad i = [2:n]. 
  \label{eq:MAc}
  \end{align}
  \end{subequations}
   
  Equation~(\ref{eq:MAa}) means that there is no ``worth'' for the empty and one-element set, because we need at least two elements to compare. 
  An example showing the calculation of the FM using Equations~(\ref{eq:MAa})--(\ref{eq:MAc}) is given in Section \ref{subsect:se}.
  
 \subsection{Object Detection Using Deep Learning in ADAS} \label{subsect:detcCa}

  In recent years, deep learning has become the focus of much research. 
  The beginning of the surge in deep learning was the winning of the ImageNet Large-Scale Visual Recognition Challenge (ILSVRC)~\citep{deng2009imagenet} by AlexNet~\citep{krizhevsky2012imagenet} developed by Alex Krizhevsky et al. Since then, deep learning has been integrated in many systems with outstanding performance, as shown in \citep{zeiler2014visualizing, simonyan2014very, szegedy2015going, he2016deep}.
  

  Due to the development of deep learning, the accuracy of object detection in images has largely been improved.
  Examples of these object detectors includes: YOLO~
	\citep{redmon2016you, redmon2016yolo9000}, Faster R-CNN \citep{ren2015faster}, Multibox \citep{szegedy2014scalable}, Region-based Fully Convolutional Networks (R-FCN) 
	\citep{li2016r} and the Single Shot multibox Detector (SSD) 
	\citep{liu2016ssd}.
  Furthermore, there are various methods focusing on improving localization accuracy. One approach utilizes a probabilistic method. One example of this approach is called an object Localization Network (LocNet)~
	\citep{gidaris2016locnet}, which relies on assigning conditional probabilities to rows and columns of the bounding box.
  LocNet exceeds the performance of Fast R-CNN \citep{girshick2015fast}, but it has been surpassed by Faster R-CNN~\citep{ren2015faster}. 
  A second approach focuses on improvement of Non-Maximum Suppression (NMS)~\citep{felzenszwalb2010object}. One example is the so--called Gossip Net (Gnet) 
	\citep{hosang2017learning}, which incorporates NMS into the detection network, but requires a large number of training data. In another example, the authors gave slight modification to NMS, but the improvement typically only happens in some specific cases~\citep{bodla2017improving}. A third type of approach uses neural networks to refine the localization. One~example is RefineNet \citep{lin2017refinenet}, which iteratively pools features from previous predictions. However, its performance using deeper networks is yet to be studied. Our~approach is different from these three~types. We use a computationally-intelligent method to fuse different detection results obtained from augmented~images.
  
  For ADAS, two of the most important metrics for the object detection sub-system are accuracy and speed. 
  When using ADAS to help a driver in the process of driving, we first need the system to accurately detect objects near or far, big or small in various environmental conditions. Misdetection or false detection could possibly lead to disasters.
  Second, the detection needs to be in real time. 
  In~our application for an industrial collision avoidance system, the maximum speed of the vehicle is about 5 m per second, which is roughly 18 km/h. To guarantee a safe stop for collision avoidance, the maximum computational time is 0.2 s, so real time in our application is defined as having a detection rate of 5 Hz or above. Other systems with different parameters will have different requirements.

 \subsection{Image Augmentation} 
 \label{subsect:imgEnh}
 
 In many deep learning systems, there is a large number of parameters to be estimated, and thus, a large amount of training data is required. Often, augmentation techniques such as image translation, rotation, stretching, shearing, rescaling, etc., are utilized to provide augmented imagery for training, and these image augmentation techniques during training often help the system be more robust. 
 After performing a thorough literature search, we found that there were rarely any previously published systems that utilized multiple augmentations and fusion for processing test images. One system is Howard's \citep{howard2013some} entry for ILSVRC \citep{deng2009imagenet} in 2013, in which he modified the winning 2012 entry \citep{krizhevsky2012imagenet} by adding augmentations in both training and testing. The training augmentations were (1) extended pixel crops, (2) horizontal flips and (3) manipulating the contrast, brightness and color. For~testing, a combination of 5 translations, 2 flips, 3 scales and 3 views yielded 90 enhanced images. A greedy 
 algorithm then down-selected this to the top ten predictions, which were combined to form the final result.
 However, this augmentation employed during testing \citep{howard2013some} yielded an increased classification accuracy, but did not improve detection localization accuracy.

\newpage
 \subsection{Model Ensembles} 
 \label{subsect:ensemble}
 {In order to improve performance during testing and add some robustness to the system, one idea is to utilize ensembles of multiple models. In this approach, one would train multiple models and average their predictions during the test time. One example is in a YOLO paper} \citep{redmon2016you}, {where the authors combined Faster R-CNN and YOLO together and obtained some performance improvement.}
 In \citep{huang2017snapshot}, {rather than training separate models independently, the authors kept multiple snapshots of the model during training. At test time, they averaged the predictions of these multiple snapshots.}
 
Malisiewicz et al.~\citep{malisiewicz2011ensemble} {trained a linear SVM for each class in the training set. The main disadvantage is each class requires millions of negative examples to train. Szegedy et al.} \citep{szegedy2014scalable} demonstrated that proposal generation can be learned from data (vs. hand-crafted) and utilized an ensemble approach to aggregate detections. Their solution was limited to one object per grid entry with a course grid, vs. YOLO, which has a finer grid and allows multiple objects to be detected per grid location. Maree et al. \citep{maree2005random} {utilized ensembles of extremely randomized decision trees and randomly extracted subwindows for object detection. The tree sizes were very large, and this method may not be appropriate for a real-time system. There are also unsupervised methods such as applying an ensemble clustering to extract ROIs from low depth of field images by Rafiee et al.} \citep{rafiee2013region}. {Performance improvements could be obtained if this were a supervised method.}
 
 {There are two major differences between the model ensemble approach and our proposed method. First, instead of using multiple models, we use one model, and we change the inputs of the model instead of the model itself. Second, instead of averaging all predictions, we propose a computationally-intelligent method to fuse all predictions together.} 
 
\section{Proposed System} 
\label{sect:methods}
\vspace{-6pt}

  \subsection{Overview}
  For the proposed system, during the in-line (testing) phase, the input goes through three stages. First, the input is augmented to produce several variations, so we can have augmented inputs for future stages. 
  Second, a detector is applied to obtain AABBs and related labels in each augmented image. Practically, this would be implemented by applying multiple detectors in parallel, one for each of the augmented images.
  This system can handle multiple objects in the input image. The~largest number of detected objects in the augmented inputs is chosen as the number of objects ($S$) in the input. For example, suppose three augmented images provide 3, 2 and 2 detections; we~then choose the conservative approach and set $S=3$. All AABBs are grouped for each object using $k$-means clustering~\citep{lloyd1982least}, where $k$ equals $S$. 
  Third, for each object, a certain number of AABBs with the top $T$ confidence scores in augmented inputs is selected, where $T$ can be any integer that is equal to or smaller than the number of augmented inputs ($M$). Then, the AABBFI fusion method is used to fuse the $T$ AABBs to obtain one AABB for each object in the input. 
  In the ADAS examples in Section \ref{sect:results}, we~use three AABBs for tractability, but more AABBs can be fused using the method proposed in \citep{islam2017data}. 
  
\begin{figure}[H]
  \centering
  \includegraphics[scale=0.63]{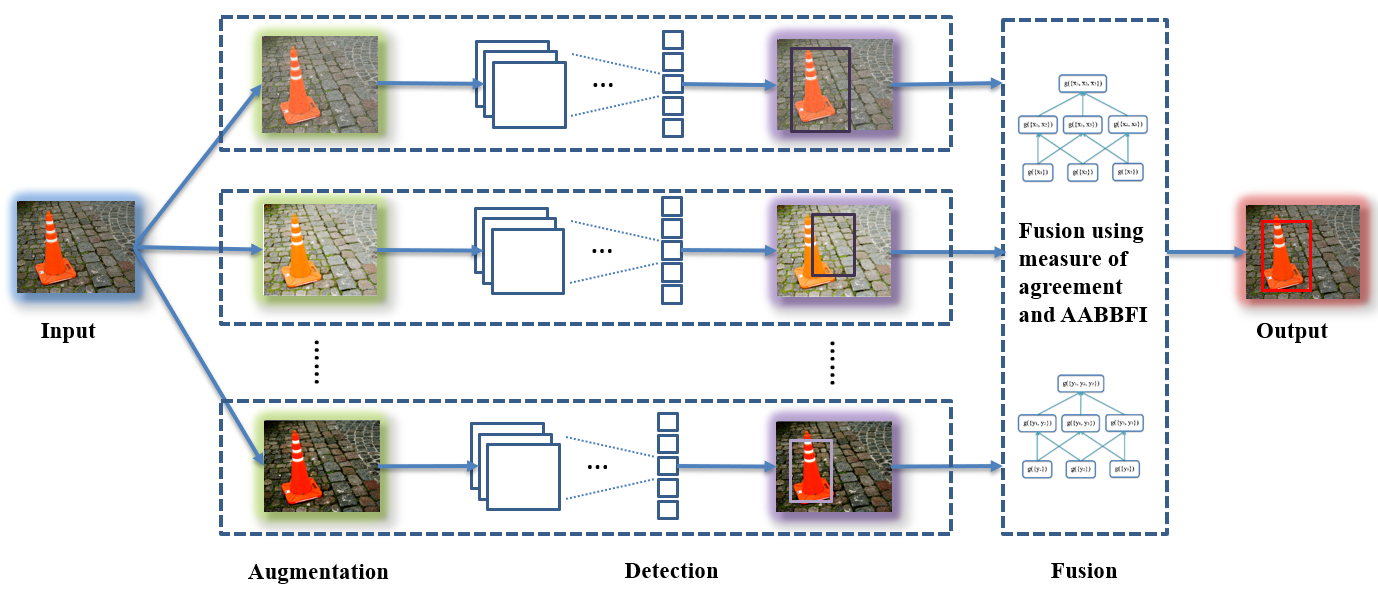}
  \caption{Proposed system for detection fusion.}
  \label{fig:prop}
  \end{figure}

  In summary, the system produces variations of each input, detects objects in each variation and fuses the top $T$ results for each object, with the expectation of getting a more accurate AABB for each object in the input. The overview of the proposed system is shown in Figure~\ref{fig:prop}, and Algorithm \ref{alg:system} is a formal description of the proposed system. 
  
   In the following paragraphs, each part of the proposed system, including augmentation, detection and fusion, is discussed in detail. 
  
\unskip

  \begin{algorithm}[H]
	\fontsize{10pt}{10pt}\selectfont
	\caption{Algorithm for the proposed detection fusion system.}
	\begin{algorithmic}[l] 
		\State \textbf{input}: input: $image \, I$ \\
		\State \textbf{output}: fused AABBs for S 
		objects: $\{\hat{B}_1, \hat{B}_2,...,\hat{B}_S\}$ \\
		\State \textbf{global}: augmentation methods: $\{P_1, P_2,...,P_M\}$ \\
		\State \textbf{global}: augmented inputs: $\{\hat{I}_1, \hat{I}_2,...,\hat{I}_M\}$\\
		\State \textbf{global}: number of objects in input: $S$\\
		\State \textbf{global}: AABBs for all objects: $\{B_1, B_2,...,B_X\}$\\
		\State \textbf{global}: labels for all objects: $\{l_1, l_2,...,l_X\}$	\\
		
		\State \textbf{global}: AABBs for each object: $\{B_1, B_2,...,B_N\}$, $N$ may vary\\
		\State \textbf{global}: labels for each object: $\{l_1, l_2,...,l_N\}$, $N$ may vary\\
		\State \textbf{global}: Number of AABBs to be fused: $T$
		\\
		\item[]
		\State \textbf{start:}
		 \\
	  \For {each augmentation method in $\{P_1, P_2,...,P_M\}$ }\\
	   \State Use the augmentation method on $I$ to obtain augmented inputs $\{\hat{I}_1, \hat{I}_2,...,\hat{I}_M\}$.\\
	  \EndFor
   \\
   \item[]
	  \State Use the detector to get $\{B_1, B_2,...,B_X\}$ and $\{l_1, l_2,...,l_X\}$ for all objects of all augmented inputs. 
   \\
   \item[]
	  \State Choose the largest number of detection in each variation as $S$. 
	  \\
	  \item[]
	  \State Group the detection into $S$ object groups using $k$-means clustering \citep{lloyd1982least}, where $k$ equals $S$.
   \\
   \item[]
	  \For {$k$ from 1 to $S$}\\
	   \State Choose the grouped detection results $\{B_1, B_2,...,B_N\}$ and $\{l_1, l_2,...,l_N\}$ for each object, where these $N$ detection results from $N$ augmented images correspond to the same object. Here, $N$ may vary.\\
     \State Choose the majority in $\{l_1, l_2,...,l_N\}$ as the object label.\\
     \If {$N \geq 3$}\\
       \State Fuse $T$ AABBs with the top $T$ confidence scores using AABBFI to get $\hat{B}_k$\\
     \ElsIf {$N = 2$}.\\
       \State Average the AABBs to get $\hat{B}_k$.\\
     \ElsIf {$N = 1$}\\
       \State Use only this detection AABB as $\hat{B}_k$.\\
     \ElsIf {$N = 0$}\\
       \State $\hat{B}_k = None$.\\
     \EndIf\\
   \EndFor
	\\
	\item[]
	\State \textbf{return} $\{\hat{B}_1, \hat{B}_2,...,\hat{B}_S\}$.
 	\end{algorithmic}
	\label{alg:system}
 \end{algorithm}

  \subsection{Augmentation}
  
  For augmentation of each input, the goal is to produce a range of inputs. The ``optimal'' augmentation cannot be determined algorithmically, and it varies depending on the type of objects detected, the image background, etc. Instead, a range of images of varying quality is generated and presented to the detector. 
  The augmentation methods used herein include changing brightness, contrast, edge enhancement, global histogram equalization, Gaussian blurring and adding independent and identically distributed (IID) Gaussian noise to simulate different scenarios. 
 \textls[-15]{Although there are other augmentation methods, in this initial work, we choose to focus on basic operations, which we have already shown lead to success. 
  Future investigations can study other more complicated local and global~operators.} 
  
  In order to see whether all these basic augmentation methods work for the final result, we choose 17 of them to test on 572 images of traffic cones and summarize the choice of each augmentation method. During testing, each augmentation method that is chosen by the proposed algorithm as the top three choices is tallied. Detection algorithms usually output confidence scores (as in YOLO~\citep{redmon2016you, redmon2016yolo9000}) or objectness scores (as in Faster R-CNN \citep{ren2015faster}) for each detected object. Herein, we use these scores to determine the top choices.

\begin{figure}[H]
  \centering
   \includegraphics[scale=1.]{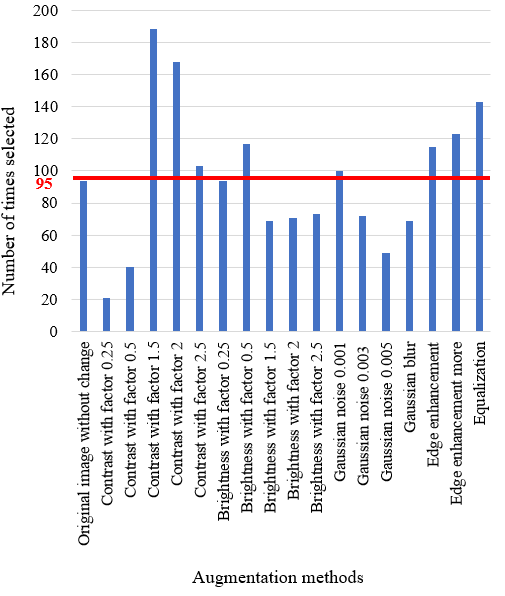}
   \caption{Augmentation analysis.} 
  \label{fig:prepro}
   \end{figure}

  We run our system and count how many times each enhancement method is chosen. The results are as shown in Figure~\ref{fig:prepro}. The number of times selected for each augmentation method is shown on the $x$-axis. The different augmentation methods are listed on the $y$-axis, including the parameters for each method. For example, ``brightness with factor 0.25'' means the image is darker than the original image, while ``brightness with factor 2'' is two times brighter than the original image.
  
  From Figure~\ref{fig:prepro}, we can see that each augmentation method is chosen in certain scenarios as one of the top three choices. For each method, if it is chosen equally, the average number of chosen times would be around $95$, as shown in the figure as a red line. In Figure~\ref{fig:prepro}, the highest number of chosen times is $188$, when using contrast with factor 1.5. The lowest chosen times is $21$, when using contrast with factor 0.25. However, all these numbers fluctuate around the average $95$, which means that each augmentation method has a certain influence on the final result, which cannot be ignored. This figure also shows that there is really no overall ``magic-bullet'' method that universally helps for all images and the detection of all objects.

\begin{figure}[H]
  \centering
   \includegraphics[scale=1.12]{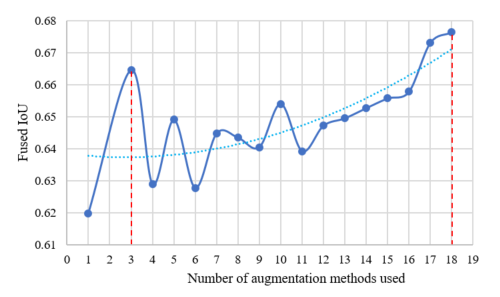}
  \caption{Changes of performance with the increase of the number of augmentation methods.} 
  \label{fig:trend}
   \end{figure}

  As the computational time scales linearly with the number of augmentation methods, we want to evaluate how the performance of the fusion system changes as the number of augmentation methods increases. In order to do this, the methods shown in Figure~\ref{fig:prepro} are sorted in descending order based on the number of times each was selected during training. The original image without any augmentation is used as a starting point. For $M$ augmented images, the $M$ methods that were selected the most often are utilized. For instance, when using $M = 3$ augmented images, the contrast with factor 1.5 and~contrast with factor 2 methods are chosen. Together with the original image, we can have three input images. The results are shown in Figure~\ref{fig:trend}.
  From this figure, we can see that the performance reaches the highest point when we choose all 18 augmentation methods. However,~the results also show fluctuations between one and 11 methods. It is important to note that when the number of augmentation methods is three, the performance reaches a local maximum point. This means that although the general trend line is going up, when computational resources are limited or there is higher speed requirement, we can choose three methods instead of 18 and still get improved performance.
  In~Figure~\ref{fig:trend}, the number of augmented methods is varied, and the IoU fusion result is calculated. The fusion really only occurs when $M \geq 3$ methods are used. The dashed vertical lines show the limits of what was examined in this paper, the $M=3$ to $M=18$ cases. The dashed blue line shows a trend line fitted to the data. The red dashed lines show the minimum (3) and maximum (18) number of augmentation methods investigated. The solid blue line shows the data trend (the data points are blue~dots).

  \subsection{Detection} 

  There are basically two types of detection (localization) methods using deep learning. One type is based on region proposals, and one example is Faster R-CNN \citep{ren2015faster}. For Faster R-CNN, the input image first runs through some convolutional layers to obtain a feature map representing the entire image, then there is a separate region proposal network, which works on top of the convolutional feature map and predicts its region proposals. 
  Once we have these predicted region proposals, we~pass them up to the rest of the network to get a classification of these regions. 
  Therefore, there are two parts in the network; one is for proposing regions, and the other is for classification. 

  The other type of detection method detects without proposing possible regions. Instead, it~localizes and classifies objects using one giant convolutional network. One typical example for this type of detector is called YOLO \citep{redmon2016you, redmon2016yolo9000}. 
  Given an input image, YOLO first divides it into some coarse grids. Within each of these grid cells, YOLO assigns some base bounding boxes. For each of these base bounding boxes, YOLO predicts an offset off the base bounding box to predict what is the true location of the object off this base bounding box, a confidence score for the location and classification scores. To summarize, YOLO uses a single neural network for simultaneously predicting the location of the object, the confidence score for the location and classification scores. 
  
  Compared with methods based on region proposals, YOLO is considered to be real time, but it also has relatively high detection accuracy. One version of YOLO training on the PASCAL VOC dataset~\citep{everingham2010pascal} can achieve 40 Frames Per Second (FPS) and 78.6 mean Average Precision (mAP), while with the same training dataset, Faster R-CNN obtains 70 mAP and 0.5 FPS. In the YOLO paper \citep{redmon2016you}, the authors admitted that they struggled to localize small objects, which can be mitigated by training special detectors for small objects, as shown in \citep{hu2017finding}. 
  
  One thing we observe from the experiment is that YOLO is more sensitive to input changes than Faster R-CNN, which means that there would be more variety in YOLO's outputs. This variety is a good fit for our proposed AABBFI method to produce a more accurate result. 
  Therefore, YOLOv2 \citep{redmon2016yolo9000} is chosen in our proposed system to achieve real-time accurate detection for robust ADAS. 
  
  \subsection{Fusion}
  
  For the fusion part, the FM of agreement as defined in Section \ref{subsec:fma} is used. Herein, the AABB Fuzzy Integral (AABBFI) is proposed to fuse AABBs.

  In this paper, evidence is represented as an AABB. In other words, evidence is an AABB indicating where the network delineated the detected object's boundaries. For example, an AABB can be represented by the coordinates of upper-left and bottom-right vertices, such as $[1, 2, 3, 4]$, which means that the coordinates for the upper-left vertex are $[1, 2]$, and the coordinates for the bottom-right vertex are $[3, 4]$. 
  However, the FI have only been extended to one-dimensional interval-valued information. Herein, we~extended it to two dimensions and specifically AABBs. Note, the advantages of an AABB are that (i) it is a convex and normal set and (ii) it can easily be decomposed without loss into two separate intervals, one~on the row and one on the column axis. Hereby, the AABBFI is defined as~follows:
  
  \begin{Definition} [\bf{AABB fuzzy integral}]
  Let $\overline{\overline{h_i}}$ be the AABB evidence from the $i${-th} input.
  Let $\overline{\overline{{h}_{x,i}}} \subseteq I$ be the interval-valued evidence for the $x$-axis of the $i${-th} AABB from the $i${-th} input. Furthermore, let $[\overline{\overline{{h}_{x,i}}}]^{-}$ and $[\overline{\overline{{h}_{x,i}}}]^{+}$ be the left and right endpoints of interval $\overline{\overline{{h}_{x,i}}}$. 
  In the same way, let $\overline{\overline{{h}_{y,i}}} \subseteq I$ be the interval-valued evidence from AABB's $y$-axis from the $i${-th} input, and let $[\overline{\overline{{h}_{y,i}}}]^{-}$ and $[\overline{\overline{{h}_{y,i}}}]^{+}$ be the top and bottom endpoints of interval $\overline{\overline{{h}_{y,i}}}$.
  Herein, the membership at each location of the AABB is assumed to be one. The AABBFI can be computed as~follows, 
  
  \begin{equation}
  \int\overline{\overline{h}} \circ g = [\int[\overline{\overline{h_{x}}}]^{-} \circ g , \int[\overline{\overline{h_{y}}}]^{-} \circ g , \int[\overline{\overline{h_{x}}}]^{+} \circ g, \int[\overline{\overline{h_{y}}}]^{+} \circ g].
  \label{eq:AABBFI}
  \end{equation}
  \end{Definition}
  
  Because an AABB is convex and the heights of all of our sets are one, we can disregard the third dimension (the ``value'' at each location in the AABB). These AABBs are assumed herein to be binary sets. 
  Furthermore, the axis-aligned convexity makes proving Equation~(\ref{eq:AABBFI}) trivial because it is easily decomposed; meaning each AABB can clearly be represented by the union of two closed intervals, and one can compute that as two separate individual interval-valued integrals and union the result (verifiable by the extension principle \citep{de2017extension}). 

\section{Examples and Discussion} 
\label{sect:results}

 In order to explain and validate the AABBFI fusion method, three synthetic examples are given. Then real-world ADAS examples are analyzed to reinforce the proposed system's efficacy.
 
 For comparison with the proposed fusion method, two expected value operations, average and median, are chosen. They are two of the most commonly-used methods for summarizing multiple inputs into one output, and median operation is supposed to be robust to noise. 
 
  \subsection{Synthetic Examples} 
  \label{subsect:se}
  
  The following are three synthetic examples of different scenarios, where the first example has three similar AABBs; the second example has three AABBs with one outlier; and third example has three AABBs with one extreme outlier. These three different synthetic examples show how this fusion process works and give us some intuition about why the proposed AABBFI method works better than average and median operations.

\begin{figure}[H]
  \centering
  \includegraphics[height=.35\textheight]{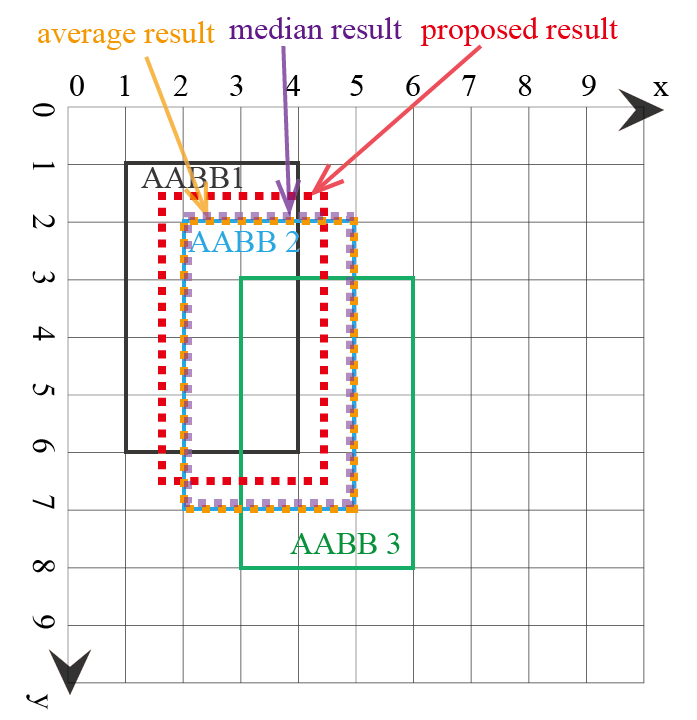}
  \caption{First example of fusing three similar AABBs.} 
  \label{fig:eg1}
  \end{figure}

\begin{figure}[H]
  \centering
  \begin{subfigure}[b]{0.459\textwidth}
  \includegraphics[width=\textwidth]{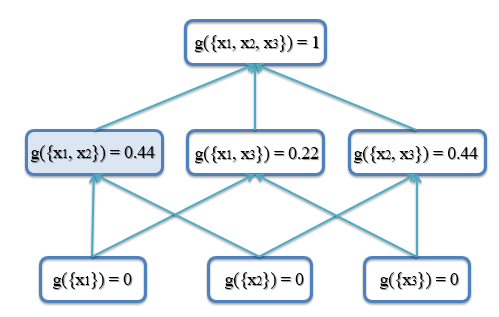}
  \caption{}
  \end{subfigure}
  \begin{subfigure}[b]{0.4\textwidth}
  \includegraphics[width=\textwidth]{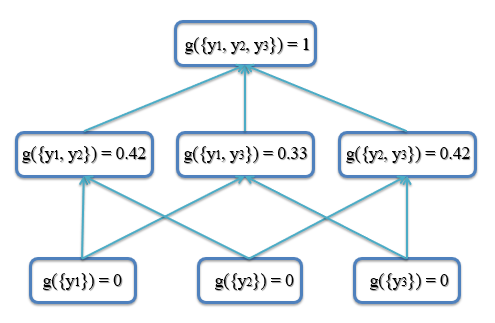}
  \caption{}
  \end{subfigure}
  \caption{Lattice of the FM of agreement for the first example of three similar AABBs. See the text for an example of how to calculate $g(\{x_1, x_2\})$. (\textbf{a}) Lattice of the FM of agreement on the x
	-axis for the first example; (\textbf{b})~lattice of the FM of agreement on the y-axis for the first example.} 
  \vspace{-12PT}

  \label{fig:fm1}
  \end{figure}

  \begin{Example}
   { Three similar AABBs:}
  \end{Example}
  In the first example, there are three AABBs to be fused. The reason these AABBs are chosen is that they have the same size, and they overlap, which means that they are similar. The top and bottom vertices' coordinates ($[x_{top}, y_{top}, x_{bottom}, y_{bottom}]$) are used to represent each AABB. The coordinates of the three AABBs are $[1, 1, 4, 6], \ [2, 2, 5, 7], \ [3, 3, 6, 8]$, as shown in Figure~\ref{fig:eg1}. Note that the $x$ and $y$ axes units are arbitrary units of measure. The lattice of the FM of agreement is shown in Figure~\ref{fig:fm1}. The result using the proposed AABBFI method is shown in Figure~\ref{fig:eg1} with the red dashed line, while the average result is shown with the yellow dashed line, and the median result is shown with the purple dashed line for comparison. The coordinates of the AABB obtained from the proposed AABBFI method are $[1.44, 1.42, 4.44, 6.42]$; the average AABB's coordinates are $[2, 2, 5, 7]$; and~the median AABB's coordinates are $[2, 2, 5, 7]$.

   \unskip

  For the calculation of the FM of agreement, take the calculation of $g(\{x_{1}, x_2\})$ as an example. For the $x$-axis, the three intervals to be fused are $x_1 =[1, 4], x_2=[2, 5]$ and $x_3=[3, 6]$. From Equation~(\ref{eq:MAb}), 
  we~have:
  \[
  \tilde{g}^{AG}(\{x_1, x_{2}\}) = \bigg|[1, 4] \cap [2, 5] \bigg| \times \frac{2}{3} = \bigg|[2, 4] \bigg| \times \frac{2}{3} = 2 \times \frac{2}{3}= \frac{4}{3}
  \]
  \vspace{-18pt}

 \begin{align*}
  \tilde{g}^{AG}(\{x_1, x_{2}, x_3\}) &= \bigg|([1, 4] \cap [2, 5]) \cup ([1, 4] \cap [3, 6]) \cup ([2, 5] \cap [3, 6]) \bigg| \times \frac{2}{3} + \bigg|[1, 4] \cap [2, 5] \cap [3, 6] \bigg| \times \frac{3}{3} \\
  &= \bigg|[2, 4] \cup [3, 4] \cup [3, 5] \bigg| \times \frac{2}{3} + \bigg|[3, 4] \bigg| \times \frac{3}{3} \\
  &= \bigg|[2, 5]\bigg| \times \frac{2}{3} + 1 \times \frac{3}{3}= 3 \times \frac{2}{3} + 1 = 3
  \end{align*}
  
  From Equation~(\ref{eq:MAc}), the calculation is: 
  \[
  {g}^{AG}(\{x_1, x_{2}\}) = \frac{\frac{4}{3}}{3} = \frac{4}{9} \approx 0.44. 
  \]
  
  This value is shown in the shaded box in Figure \ref{fig:fm1}a.
 
  Next, we show how to calculate the $x_{top}$ coordinate using the proposed method.
  The $x_{top}$ coordinates from the three AABBs are 1, 2 and 3, which means that the permutation sorting gives $[{\overline{\overline h}}(x_{\pi(1)})]^{-} = 3, [{\overline{\overline h}}(x_{\pi(2)})]^{-} = 2, [{\overline{\overline h}}(x_{\pi(3)})]^{-} = 1$. By using the FM in Figure~\ref{fig:fm1}a, we can calculate the $x_{top}$ using Equations~(\ref{eq:CFI}) and~(\ref{eq:AABBFI}) as follows: 
  \begin{equation*}
  \small
  \begin{array}{cl}
   x_{top} = \int[\overline{\overline{h_{x}}}]^{-} \circ g &= 3 \times[g(\{x_{3}\}) - 0] + 2 \times[g(\{x_{2}, x_3\}) - g(\{x_{3}\})] + 1 \times[g(\{x_1, x_{2}, x_3\}) - g(\{x_{2}, x_3\})]\\
  &= 3 \times (0 -0) + 2 \times (0.44 -0) + 1 \times (1 - 0.44) = 1.44.
  \end{array}
 \end{equation*}

  In this example, three AABBs are similar, with the same size and overlapping with each other. Therefore, we expect the fused AABB to be influenced by all three AABBs, both in shape and position. From the result, we can see that the majority of the proposed AABB is the overlapping area of the three AABBs, and it has the same size, just like our expectation. On the other hand, the average AABB and median AABB are exactly the same as AABB 2, which may be or may not be influenced by AABB 1 and AABB 3. In the case of fusing three similar AABBs, the proposed method gives a result that fits our natural expectations. 
  
\begin{figure}[H]
  \centering
  \includegraphics[height=.3\textheight]{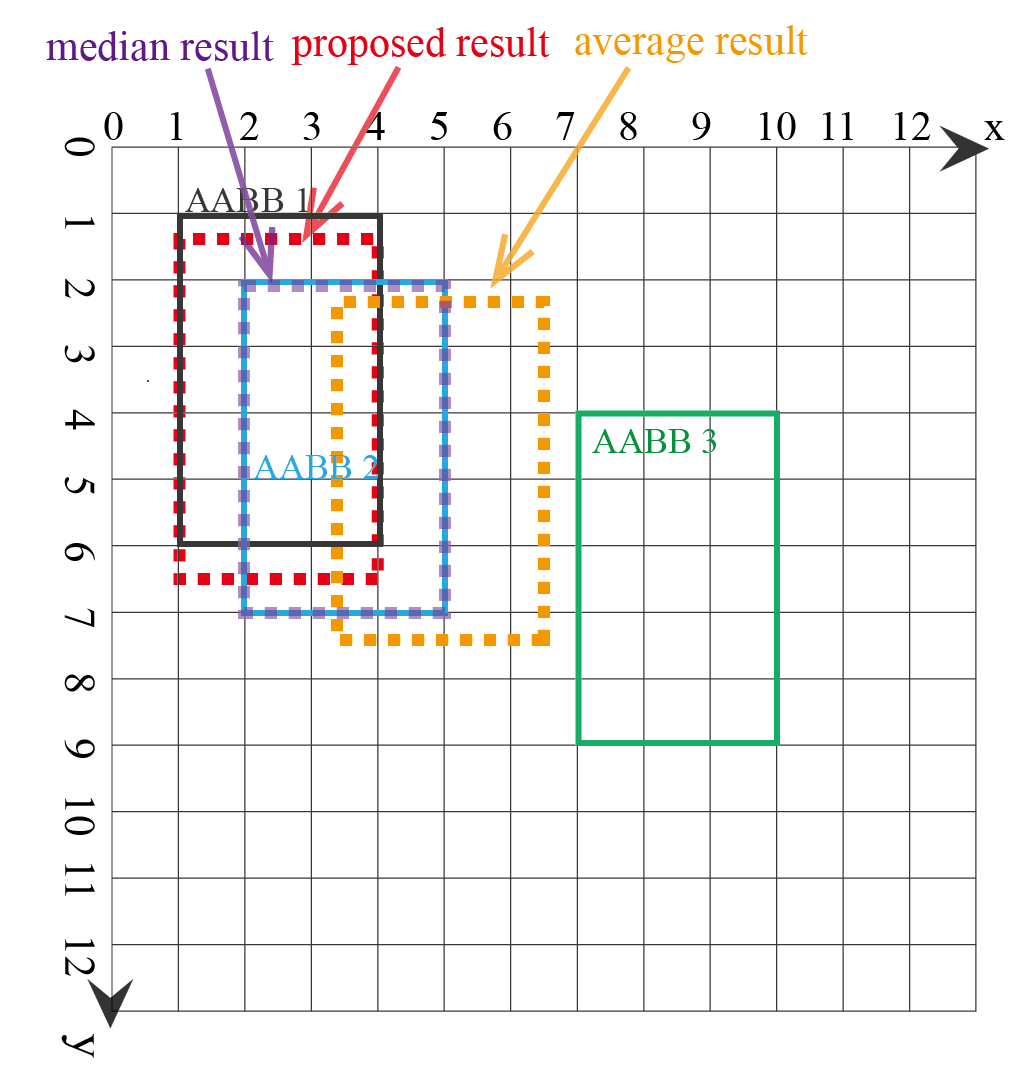}
  \caption{Second example of fusing three AABBs with one outlier.} 
  \label{fig:eg2}
  \end{figure}
\begin{figure}[H]
  \centering
  \begin{subfigure}[b]{0.47\textwidth}
  \includegraphics[width=\textwidth]{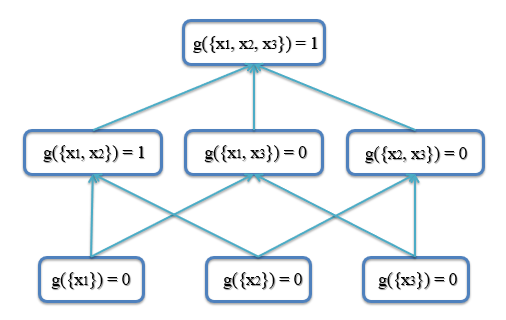}
  \caption{}
  \end{subfigure}
  \begin{subfigure}[b]{0.47\textwidth}
  \includegraphics[width=\textwidth]{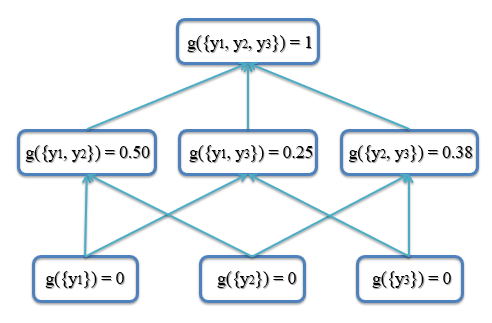}
  \caption{}
  \end{subfigure}
  \vspace{-12pt}

  \caption{Lattice of the FM of agreement for the second example of three AABBs with one outlier. (\textbf{a})~Lattice of the FM of agreement on the x-axis for the second example; (\textbf{b}) lattice of the FM of agreement on the y-axis for the second example.} 
  \label{fig:fm2}
  \end{figure}

  \begin{Example} 
  {Three AABBs with one outlier:}
  \end{Example}
 \textls[-10]{In the second example, there are three AABBs to be fused, with one possible outlier being AABB 3, as AABB 3 has no overlapping area with AABBs 1 and 2. The three AABBs are $[1, 1, 4, 6],$ $\ [2, 2, 5, 7],$  $\ [7, 4, 10, 9]$, as shown in Figure~\ref{fig:eg2}. The lattice of the FM of agreement is shown in Figure~\ref{fig:fm2}. The proposed AABBFI result is shown in Figure~\ref{fig:eg2} with the red dashed line, with the average result with the yellow dashed line and median result with the purple dashed line}. The proposed AABB's coordinates are $[1, 1.38, 4, 6.38]$; the average AABB's coordinates are $[3.33, 2.33, 6.33, 7.33]$; and the median AABB's coordinates are $[2, 2, 5, 7]$.
 
 In this example, AABB 3 has no overlap with the other two AABBs, and it seems like an outlier even though it is the same size. We expect that the fused result should have more agreement with the first two AABBs. From the result, it is shown that the AABB from the proposed AABBFI method is most similar to AABB 1 and influenced by AABB 2 without any overlap with possible outlier AABB~3. On the other hand, if averaging the three AABBs, the result is shifted towards AABB~3, which means that the average result is influenced heavily by AABB~3. The median result is again the same as AABB 2, without any influence from either AABB 1 or AABB 3. The two results by average and median operations do not fit our expectation, while the result produced by the AABBFI method fits our expectation that the fused result is largely influenced by AABB 1 and AABB 2, with little influence from AABB 3.

  \unskip
  
  
\begin{figure}[H]
  \centering
  \includegraphics[height=.31\textheight]{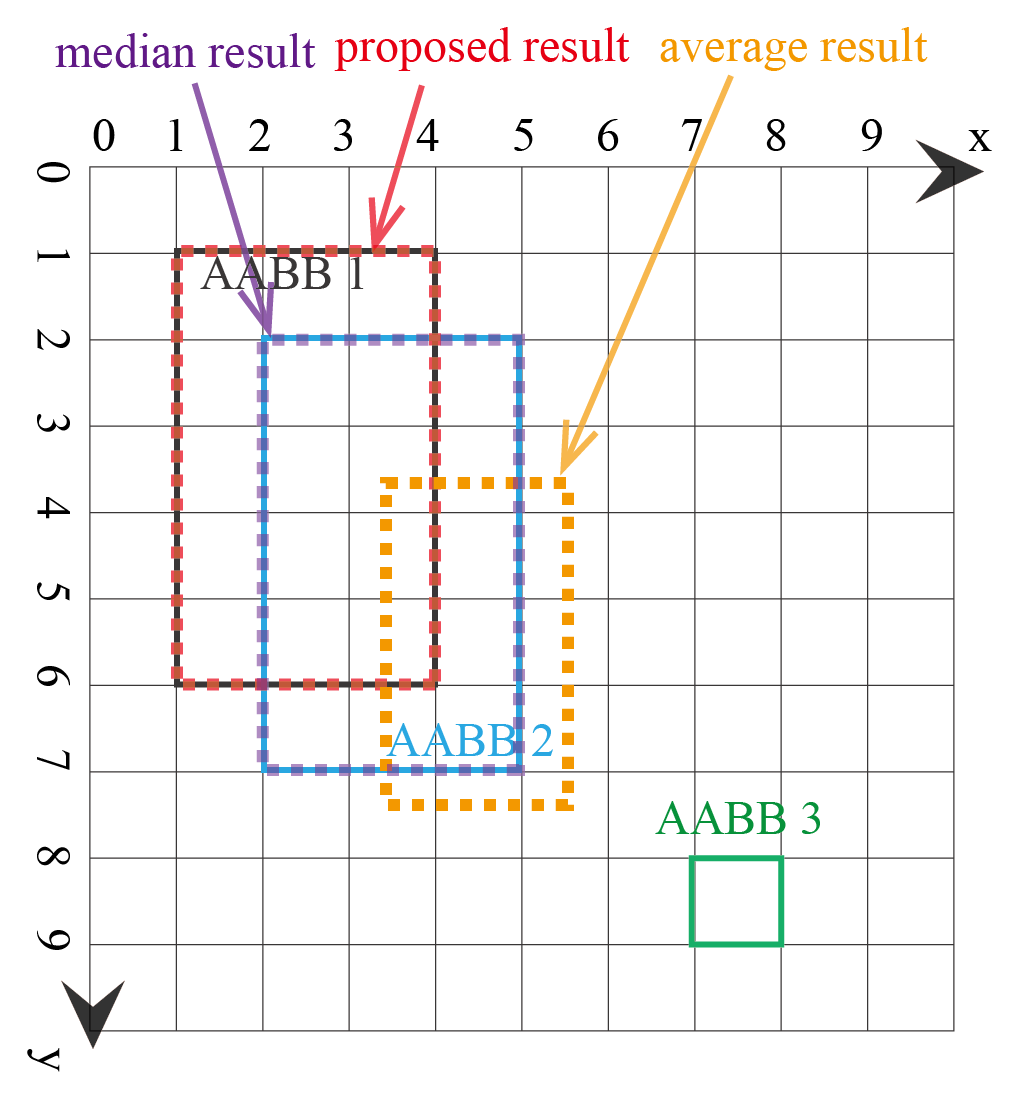}
  \caption{Third example of fusing three AABBs with one extreme outlier.} 
  \label{fig:eg3}
  \end{figure}

\begin{figure}[H]
  \centering
  \begin{subfigure}[b]{0.49\textwidth}
  \includegraphics[width=\textwidth]{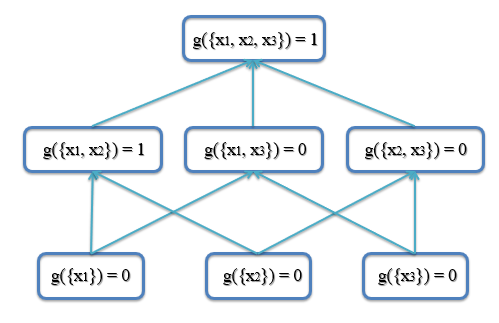}
  \caption{}
  \end{subfigure}
  \begin{subfigure}[b]{0.49\textwidth}
  \includegraphics[width=\textwidth]{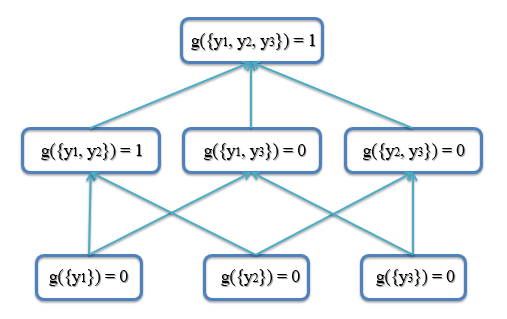}
  \caption{}
  \end{subfigure}
  \vspace{-12pt}

  \caption{Lattice of the FM of agreement for the third example of three AABBs with one extreme outlier. (\textbf{a}) Lattice of the FM of agreement on the x-axis for the third example; (\textbf{b}) lattice of the FM of agreement on the y-axis for the third example.}
  \label{fig:fm3}
  \end{figure}

  \begin{Example} 
  {Three AABBs with one extreme outlier:}
  \end{Example} 
  \textls[-5]{In the third example, there are three AABBs to be fused with one obvious outlier AABB 3, which has a much smaller size than the other two and has no overlapping area. The three AABBs are $[1, 1, 4, 6],\ [2, 2, 5, 7],\ [7, 8, 8, 9]$, as shown in Figure~\ref{fig:eg3}. The lattice of the FM of agreement is shown in Figure~\ref{fig:fm3}. The result from the proposed AABBFI method is shown} in Figure~\ref{fig:eg3} with the red dashed line. The~average result is shown with the yellow dashed line and the median result with the purple dashed line. The~proposed AABB's coordinates are $[1, 1, 4, 6]$; the average AABB's coordinates are $[3.33, 3.66, 5.67, 7.33]$; and the median AABB's coordinates are $[2, 2, 5, 7]$.

 In this example, AABB 3 looks like an obvious outlier and should not influence the final fused result at all. From the fused result using the AABBFI method, we can see that the fused AABB is the same as AABB 1, without any influence from AABB 3, which is the same as our expectation. On the other hand, the average AABB is largely influenced by AABB 3. The average AABB not only shifts towards AABB 3, but also shrinks in size, which is different from our expectation. Once again, the median result is the same as AABB 2. Comparing the three fusion results, AABB by the proposed AABBFI fits our expectation most.

  \unskip
  
  \subsection{ADAS Examples} 
  \label{subsect:adas}
  
  In the ADAS examples, we focus on the accurate detection of three types of objects: cones, pedestrians and boxes. 
  The proposed system is also tested on all of the PASCAL VOC 2007 testing dataset, and mAP is calculated to show general improvement on detection \citep{everingham2010pascal,everingham2011pascal}. 

  One metric for measuring the accuracy of an object detector is called Intersection over Union (IoU), also known as the Jaccard index \citep{jaccard1901etude}. This index is used to compute the similarity between finite sets. Suppose we have two sets $A$ and $B$, which could be two AABBs. To measure similarity between them, IoU can be used and is calculated as follows, 
  \begin{equation}
  IoU = J(A, B) = \frac{Area \ of \ Overlap}{Area \ of \ Union} = \frac{|A \bigcap B |}{|A \bigcup B |}.
  \label{eq:IoU}
  \end{equation}
  
  Herein, we use IoU as the primary metric for detection accuracy. We also utilize mAP to provide a metric that can be compared to many results in the literature.
  
  \subsubsection{Augmentation Methods}
  
  In the experiments performed, the Python Imaging Library (PIL) 
  is utilized \citep{pil} to obtain augmented images. In PIL brightness and contrast enhancement classes, a factor is used for the change of brightness and contrast. When this factor is one, it gives the original image. Based on the experimental evaluation, in~ADAS examples, factor values are chosen to be 1, 0.25, 0.5, 1.5, 2.0 and 2.5 for brightness and contrast. 
  To add Gaussian noise, the noise variance is chosen to be 0.001, 0.003 and 0.005. This is based on qualitative image assessment, since these noise levels do not drastically alter the appearance of the input image.
  Other augmentation methods, including edge enhancement, global histogram equalization and Gaussian blurring (radius = 2), are predefined image operations in PIL.
  
  \subsubsection{Datasets}
  
  The primary research task is developing a robust system for ADAS, and two of the major objects that are crucial for detection in ADAS are traffic cones and pedestrians. In our specific industrial application, we also need to detect boxes, because they are commonplace items in our project's industrial setting. Therefore, we collect cone, pedestrian and box images in controlled scenarios, which produce variations in different angles and distances. 
  Specifically, the image dataset is collected using an FLIR Chameleon3 USB camera with a Fujinon 6-mm lens. 
  Training images also include images taken by a Nikon D7000 camera and images from the PASCAL VOC dataset \citep{everingham2010pascal}. The total number of training images is around 22,000. 
  For testing images, one cone, pedestrian or box is randomly placed or stands at different locations from 5 m to 20 m away and $-$20 degrees to +20 degrees in azimuth (left to right) from the camera, which covers the FLIR camera's full field of view horizontally and the detection range. For traffic cones, there are 287 testing images; for pedestrian, there are 208 testing images; and for boxes, there are 200 testing images. For generalization purposes, we also show results for the PASCAL VOC 2007 test dataset \citep{everingham2010pascal}.

  \subsubsection{Training Parameters}
  

  
  The network structure for YOLO is basically the same as the default structure of YOLOv2 \citep{redmon2016yolo9000} except the last layer. YOLOv2 divides each image into 13 by 13 grids. For our application, we predict five base bounding boxes for each grid in the image. For each box, there are 4 numbers (for the top left and bottom right coordinates of the AABB), 1 confidence score and 5 class scores. Therefore, we change the filter size to $13 \times 13 \times 50$ ($5 \times (4 + 1 + 5) = 50$) in the last layer.
  We also change the learning rate to $10^{-5}$ to avoid divergence. We use a batch size of 64, a momentum of 0.9 and a decay of 0.0005, which are the same as those in the original YOLOv2 configuration.
  
  For Faster R-CNN, the model is trained with a VGG net. Most parameters are set to be the same as the original Faster R-CNN \citep{ren2015faster}. The changes are as follows: the number of classes is modified to five; the number of outputs in the class score layer is modified to five; the number of outputs in the bounding box prediction layer is modified to 20 ($4 \times 5$). 
  
  \subsubsection{Results} 
  \label{subsubsec:results}
  
\begin{figure}[H]
  \centering
  \includegraphics[height=.253\textheight]{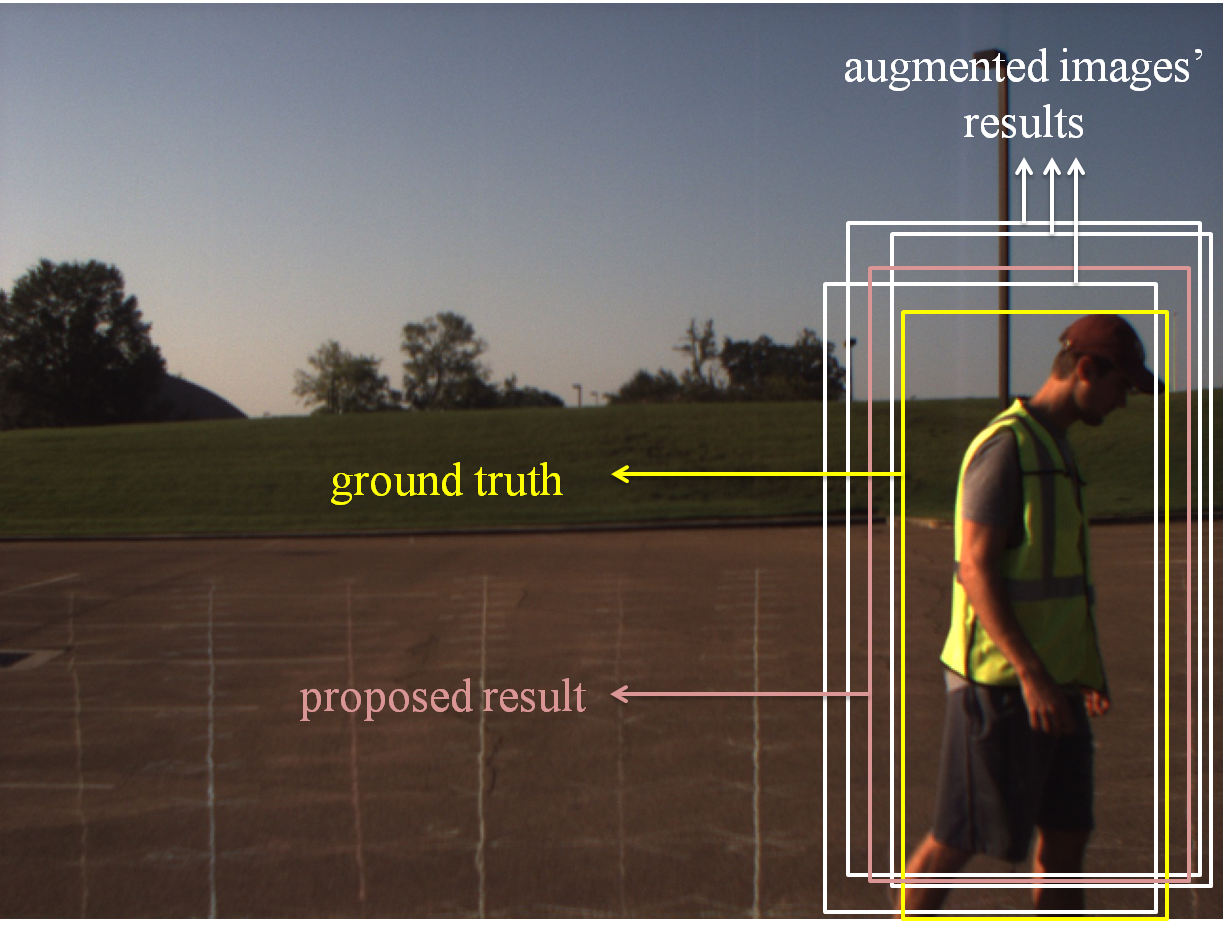}
  \caption{Fusion of three detection AABBs (white) into one AABB (pink), compared with the ground-truth (yellow).}
  \label{fig:fusion}
  \end{figure}

  Figure~\ref{fig:fusion} is one example image for pedestrian detection with ground-truth AABB, three AABBs from different augmented images and fused AABB using the proposed method. In this image, the yellow AABB is the ground-truth, which is human-determined and hand-labeled. Three augmentation methods are chosen, which are global histogram equalization, changing contrast with factor two and changing brightness with factor 2.5. The AABBs obtained from these augmented images are shown in white. The fused AABB using the proposed AABBFI method is shown in pink. 
  
\begin{figure}[H]
  \centering
  \includegraphics[height=.35\textheight]{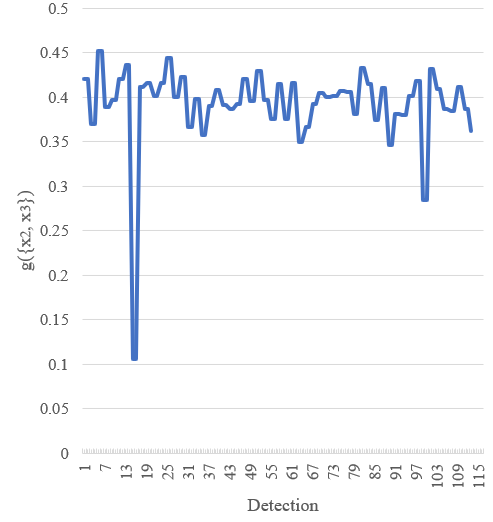}
  \caption{Value changes of $g(\{x_2, x_3\})$ for the FM of agreement on the x-axis for cone detection.}
  \label{fig:gx23}
  \end{figure}

  The value of $g(\{x_2, x_3\})$ for the FM of agreement on the $x$-axis for cone detection is shown in Figure~\ref{fig:gx23}. From this figure, we can see that the value of the FM of agreement substantially changes from case to case. In other words, for each detection, the proposed fusion method gives different ``weight'' to each AABB, unlike the average operation, which gives the same ``weight'' around $0.33$ when the number of AABBs to be fused is three.

    In the experiment, we use a PC with an Intel i5-4460 dual-core CPU and one Nvidia GTX 1080 GPU. The computational overhead for AABBFI-based fusion is around 2.15 milliseconds (466 FPS), while for processing 10 augmented inputs, it is around 142.86 milliseconds (7 FPS). The fusion overhead is small compared to the overhead for using multiple inputs. The computational load is linear with the number of given inputs. More discussion about computational complexity is given in Section \ref{sect:dis}.

  All ADAS experiment results are shown in Table~\ref{tab:compare1}. In this table, the ``No fusion'' column means that the results are detection outputs from the original images without any fusion method applied. Under the ``With fusion'' section, four fusion methods are listed: NMS, average, median and the proposed AABBFI. 

   \begin{table}[H]
   \caption{ADAS experimental results. Baseline means there is no fusion performed, that is only the original image is processed.}
   \label{tab:compare1}
  \centering
      \begin{tabular}{llllll}
   \toprule 
     & \textbf{No Fusion} & \multicolumn{4}{c}{\textbf{With Fusion}} \\ \midrule
\textbf{Fusion method} & \textbf{Baseline} &\textbf{NMS} &\textbf{Average}  & \textbf{Median}  & \textbf{Proposed}   \\
 \midrule        
   \multicolumn{6}{c}{YOLO 
	Cone}                      \\ \midrule
   \rule[-1ex]{0pt}{2.5ex}  Average IoU& 0.6199 &0.6676 & 0.6676 & 0.6717 & \textbf{0.6763}  \\
   \rule[-1ex]{0pt}{2.5ex}  Detection& 260/287 & \textbf{287/287} & \textbf{287/287} & \textbf{287/287}& \textbf{287/287}  \\ 
   \midrule
   \multicolumn{6}{c}{YOLO Box} \\ \midrule
   \rule[-1ex]{0pt}{2.5ex}  Average IoU & 0.6722 &0.6756 & 0.6925 & 0.6872 & \textbf{0.7031} \\
  
   \rule[-1ex]{0pt}{2.5ex} Detection & \textbf{200/200} & \textbf{200/200} & \textbf{200/200} & \textbf{200/200} & \textbf{200/200} \\
   \midrule
   
   
   
   \multicolumn{6}{c}{YOLO Pedestrian} \\ \midrule
   \rule[-1ex]{0pt}{2.5ex}  Average IoU & 0.7727 &0.7023 & 0.7995 & 0.7966 & \textbf{0.8141} \\
  
   \rule[-1ex]{0pt}{2.5ex} Detection & \textbf{208/208} & \textbf{208/208} & \textbf{208/208} & \textbf{208/208} & \textbf{208/208} \\
   \midrule

   \multicolumn{6}{c}{Faster R-CNN Pedestrian} \\ \midrule
   \rule[-1ex]{0pt}{2.5ex}  Average IoU & 0.7402 & 0.7377 & 0.7660 & 0.7581 & 0.7499 \\
 
   \rule[-1ex]{0pt}{2.5ex} Detection& 206/208 & 206/208 & 206/208 & 206/208 & 206/208 \\
   \midrule
   
   \multicolumn{6}{c}{YOLO VOC 2007 All} \\ \midrule
   \rule[-1ex]{0pt}{2.5ex}  Average IoU & 0.6106 & 0.6654 & 0.6678 & 0.6728 & \textbf{0.6746}  \\
   
   \rule[-1ex]{0pt}{2.5ex} mAP & 71.10\% & 72.36\% & 72.82\% & 72.41\% & \textbf{72.87\%} \\
   \bottomrule

   \end{tabular}
      \end{table}
     \subsection{Discussion} 
  \label{sect:dis}
  
   For all five fusion/non-fusion methods, the average IoU obtained and the number of detections are listed. We find that for cones, pedestrians and boxes, the highest accuracy and detection happen when using the proposed AABBFI method together with YOLO. For the PASCAL VOC dataset \citep{everingham2010pascal}, we use the threshold of 0.25 for the confidence score to calculate mAP, and we find that after fusion, not only the average IoU increases, but also mAP increases.

  In Table~\ref{tab:compare1}, we also include results on detecting pedestrians using Faster R-CNN \citep{ren2015faster}. Comparing the results, it is shown that YOLO outperforms Faster R-CNN both in accuracy and detection. Moreover, in the case that we choose Faster R-CNN as the detector, the AABBFI method does not help with accuracy compared with the average and median methods. Furthermore, Table \ref{tab:compare1} shows that mAP improves by using fusion. The mAP improved the most by using the proposed method. In table \ref{tab:compare1}, the best results are shown in bold.
  
  From the synthetic results, it is clear that with the presence of possible or obvious outliers, the~proposed AABBFI method can fuse the AABBs in a way more similar to our expectation, and it also can reduce or eliminate the effects of possible or obvious outliers. 
  
  From the ADAS experimental results, there is improvement both on IoU and detection by using the proposed AABBFI method together with YOLO. Specifically, in most cases, fusion methods help with increasing detection. It is also shown that comparing with input images' average IoU without fusion, by using the average, median and proposed fusion methods, the average IoU results increase. Comparing NMS, average, median and the proposed fusion operation, the proposed AABBFI method gives the average IoU the most improvement.
  
  From Figure~\ref{fig:gx23}, we can see that unlike the static operations such as average and median, the~proposed fusion method is dynamic. The FM of agreement substantially changes from case to case. Consequently, the fusion is done in a detection-by-detection and object-by-object manner. For fusion, we evaluate the FM of agreement among AABBs for each detection and fuse them using a computationally-intelligent method named AABBFI. As mentioned in Section \ref{sect:intro}, the fusion for detection in this scenario cannot be learned in general and is done in a case-by-case manner in this~paper.
  
  For the proposed AABBFI method, it works when the networks have variety in the results. Faster R-CNN is a region proposal-based method, which means that it uses part of its network to propose possible regions that could have objects we want to detect, and this may reduce the variety in the output. On the other hand, YOLO predicts both AABBs and labels through one giant network, which makes it fast. Furthermore, this kind of one-shot network design may produce more variations, which leads to more accurate output by using the AABBFI method. As a result, we choose YOLO together with the AABBFI method in the proposed system. The results from the ADAS experiments show that in cone, pedestrian and box detection, the proposed method gives the best results.
  
 \textls[-10]{ In the ADAS examples, the improvement of both detection and IoU could come from two aspects. First, variation can lead to detection. In the original image, the detector might not be able to detect cones, pedestrians or boxes, but in its variations, the detector may work. 
  Second, fusion based on agreement stabilizes the finally AABB. AABBFI is a dynamic fusion method that produces different result when the ``agreement'' among inputs changes. NMS is also a dynamic fusion method, but the results using it are not as good as the proposed fusion method. On the other hand, the average and median operations are both static, which means that they do not consider the specific characteristics of each input}. 
  
  To guarantee a real-time ADAS, we expect that the fusion adds as little computational complexity as possible to the original detection system. The computational complexity of the proposed AABBFI (fusion only) is actually tiny, which is $O(n)$, where $n$ stands for the number of input images. It~is the same as NMS, average and median operation. This means that the four fusion methods all have the same low computational complexity and do not add much computational burden to the system. Another thing that may slow down the system is there are multiple augmented inputs that need to be processed. 
  We can avoid the slowdown either via multiple processing units or a more powerful processing unit. This highlight a trade-off in many systems: there are multiple approaches, some~more computationally complexity, and some less computationally complexity. Given sufficient hardware resources to keep an appropriate frame rate, the proposed solution is feasible. The~computational overhead in Section \ref{subsubsec:results} shows that with ten augmented images, using our hardware configuration, we can achieve real-time performance in our definition (5 Hz or above).

  In summary, from ADAS examples, we find that AABB-based fusion with YOLO produces the highest accuracy and detection results compared with other methods listed. From the synthetic example, we could obtain some rationality behind this improvement.
  The proposed system increases the accuracy in the detection stage while maintaining the real-time characteristic in the original detection system, which makes this a realistic solution for a real-time domain like ADAS.

\section{Conclusions and Future Work} 
\label{sect:conclusionsfuture}

This paper proposed a computational intelligence system for more accurate object detection in real time. This system uses augmentation methods before a deep learning detector and then an AABB fuzzy integral (AABBFI) on the resultant AABBs. Three synthetic examples show the rationality of using the AABBFI rather than NMS/average/median operation. Experimental ADAS examples show that for real-world datasets, the proposed system gives the highest accuracy and detection results, without adding much computational complexity for real-time purposes. Our proposed system is not only fast, but also accurate, which are two important criteria in ADAS. By using this computational intelligence system, we are able to build a more robust object detection sub-system for ADAS applications, with the proposed system showing improvement in both IoU and mAP metrics. Furthermore, very good results were obtained when only utilizing three combined inputs, making the computational load roughly three-times the load for only using one input (the original image).

In the future, our next planned effort includes: (i) exploring different deep learning-based methods, (ii) studying why deep learning-based methods have such sensitivity to input variations, (iii) incorporating the detector's confidence score in the detection result as height, (iv) non-AABB region fusion (non-convex and non-normal) and (v) using various FMs beyond the FM of agreement.

Finally, we wish to thank the anonymous reviewers for their constructive comments, which have strengthened our paper.


\vspace{6pt} 
\authorcontributions{Pan Wei conceived the proposed algorithm, designed and executed experiments, and wrote a majority of the paper. John Ball and Derek Anderson reviewed experimental results, provided technical guidance, reviewed and authored parts of the paper.} 

\conflictsofinterest{The authors declare no conflict of interest.} 

 \reftitle{References}

\end{document}